\documentclass{article}

\usepackage{microtype}
\usepackage{graphicx}
\usepackage{subfigure}
\usepackage{booktabs} 

\usepackage{hyperref}

\usepackage[accepted]{icml2021}

\icmltitlerunning{Statistical Privacy Guarantees of Machine Learning Preprocessing Techniques}

\begin{document}

\twocolumn[
\icmltitle{Statistical Privacy Guarantees of Machine Learning Preprocessing Techniques}




\begin{icmlauthorlist}
\icmlauthor{Ashly Lau}{ic}
\icmlauthor{Jonathan Passerat-Palmbach}{ic,ch}
\end{icmlauthorlist}

\icmlaffiliation{ic}{Department of Computing, Imperial College London, London, United Kingdom}
\icmlaffiliation{ch}{ConsenSys Health}

\icmlcorrespondingauthor{Ashly Lau}{ashly.lau17@imperial.ac.uk}
\icmlcorrespondingauthor{Jonathan Passerat-Palmbach}{j.passerat-palmbach@imperial.ac.uk}

\icmlkeywords{Machine Learning, Differential Privacy, TPDP, Preprocessing}

\vskip 0.1in
]



\printAffiliationsAndNotice{\icmlEqualContribution} 

\begin{abstract}
\vskip 0.1em
Differential privacy provides strong privacy guarantees for machine learning applications. Much recent work has been focused on developing differentially private models, however there has been a gap in other stages of the machine learning pipeline, in particular during the preprocessing phase. Our contributions are twofold: we adapt a privacy violation detection framework based on statistical methods to empirically measure privacy levels of machine learning pipelines, and apply the newly created framework to show that resampling techniques used when dealing with imbalanced datasets cause the resultant model to leak more privacy. These results highlight the need for developing private preprocessing techniques.
\end{abstract}

\vspace{-0.5cm}
\section{Introduction}

Differential Privacy (DP) is now the standard definition for privacy, and has been quickly embraced by researchers in the Machine Learning (ML) community, with much of its recent focus being on developing differentially private optimizers like DP-SGD \cite{abadi_deep_2016}. One crucial aspect to training good machine learning models is the quality of the preprocessing pipeline. However there is a gap in literature surrounding differentially private data preprocessing techniques for machine learning applications. Ji et al. \yrcite{ji_differential_2014} state that "if the rule in preprocessing is independent of the private dataset, the preprocessing is privacy-free", which may suggest why it is overlooked. However in many cases, the choice of performing data preprocessing and its design is made from observations of the data in itself, presenting a case for measuring the effects of such preprocessing methods on differential privacy guarantees.

\subsection{Contributions}

We build on the privacy violation detection framework introduced by \cite{ding_detecting_2018}. By adapting the framework for machine learning contexts and running the framework at different levels of granularity, we can use the counterexample generator to measure and compare privacy levels of different machine learning pipelines. Specifically, this presents a novel way of measuring DP levels for machine learning algorithms via a statistical approach, whereas previous works \cite{nasr_adversary_2021, jagielski_auditing_2020} have typically done so through adversarial approaches which are relatively involved to set up.


Through this technique, we show that we can detect privacy leakage from access to just a few model outputs alone, and that resampling techniques performed as a preprocessing step in machine learning pipelines cause the resultant model to leak more privacy than without. 

\vskip 0.1em
\section{Background and Related Work}

\subsection{Differential Privacy}

Differential privacy, as introduced by Dwork et al. \yrcite{dwork_algorithmic_2013}, constitutes a strong standard for privacy guarantees for algorithms. An algorithm $\mathcal{M}: \mathcal{D} \to \mathcal{R}$ with domain $\mathcal{D}$ and range $\mathcal{R}$ satisfies $(\epsilon, 0)$-differential privacy if for any two adjacent datasets $D_1, D_2 \in \mathcal{D}$ it holds that $P(M(D_1) \in E) \le e^\epsilon * P(M(D_2) \in E)$.

The notion of group privacy follows from the definition above, where any $(\epsilon, 0)$-DP mechanism $\mathcal{M}$ is $(k\epsilon, 0)$-DP for groups of size $k$. For adjacent datasets $D_1$ and $D_2$ differing by at most one group of size $k$, we have $P(M(D_1) \in E) \le e^{k\epsilon} * P(M(D_2) \in E)$. We utilise this definition when constructing adjacent datasets.

Other useful properties include the composition theorem and the post-processing theorem. The former states that independent composition of multiple DP mechanisms is still differentially private, through adding up the components’ respective privacy budgets. Whereas the post-processing theorem guarantees that the output of any DP mechanism can be arbitrarily post-processed while maintaining its original differential privacy guarantees.

\subsection{Differentially Private Machine Learning (DPML)}

The nature of machine learning models poses a risk of them exposing information about sensitive training data. Attacks such as membership inference \cite{shokri_membership_2017} and de-anonymisation \cite{orekondy_gradient-leaks_2018} highlight just how much information can be gleaned from machine learning models about the training dataset.

Differentially private designs of machine learning models achieve privacy through adding a calibrated amount of noise to the model or its outputs. There are DP implementations for traditional machine learning (eg. Naive Bayes \cite{vaidya_differentially_2013}, Linear Regression \cite{sheffet_private_2015}), as well as for deep learning (eg. DP-SGD \cite{abadi_deep_2016}).

\subsection{Effect of Dataset Imbalance}

Imbalance in datasets present significant challenges to machine learning, much more so with the inclusion of differential privacy. Nasr et al. \yrcite{nasr_adversary_2021} reveal that DP exacerbates existing biases in data and has a disparate impact on accuracies of different subgroups of data.

Methods used to combat data imbalance include resampling techniques and the generation of synthetic datasets. However, there is not much understanding of the effects of such techniques on the DP guarantees of the resultant model.

\subsection{Measuring Differential Privacy}

Despite being widely used in practical applications, concepts surrounding DP can be easily misunderstood, and DP algorithms are difficult to implement correctly. It is important to develop auditing techniques to verify that these algorithms do not leak more privacy than stated.

\subsubsection{Auditing DPML}

In our analysis of recent literature, attempts to audit the privacy guarantees of DPML algorithms have been done through adversarial instantiation \cite{nasr_adversary_2021}. That is, simulating an adversarial attack by creating an adversary to predict whether a model was trained on a dataset $D$ or an adjacent dataset $D'$. Through analysing how well the adversary can distinguish between models trained on the two datasets, one can determine the level of privacy that the training algorithm can afford.

Such adversaries are developed with varying capabilities including access to intermediate model parameters, and ability to poison datasets and model gradients.

\subsubsection{StatDP}
\label{statdp}

\vspace{-0.2cm}
\begin{algorithm}[h]
  \caption{Overview of Counter Example Generator}
  \label{alg:counterexample-detector}
\begin{algorithmic}
  \STATE {\bfseries Input:} mechanism $M$, desired privacy $\epsilon$
   
  \STATE // Get list of possible inputs: $(D_1, D_2, args)$
  
  \STATE InputList $\longleftarrow$ InputGenerator($M, \epsilon$)
   
  \STATE $E, D_1, D_2, args$ $\longleftarrow$ EventSelector($M, \epsilon$, InputList)
   
  \STATE $\rho_{\top}, \rho_{\bot}$ $\longleftarrow$ HypothesisTest($M, \epsilon, D_1, D_2, args, E$)
   
  \textbf{return} $\rho_{\top}, \rho_{\bot}$
   
\end{algorithmic}
\end{algorithm}

StatDP is a framework proposed by Ding et al. \yrcite{ding_detecting_2018} that detects violations of DP through an empirical approach. For a mechanism $\mathcal{M}$ that does not satisfy $\epsilon$-DP, the goal is to prove this failure, i.e., to find $D_1, D_2, E$ such that $P(M(D_1) \in E) >  e^\epsilon * P(M(D_2) \in E)$.

This is estimated by repeatedly running $M(D_1)$ and $M(D_2)$ and counting the number of times the output falls into $E$. A statistical test is then used to reject or fail to reject the null hypothesis $P(M(D_1) \in E) \le  e^\epsilon * P(M(D_2) \in E)$. The hypothesis test is performed for each $\epsilon$ close to the expected $\epsilon_0$, for several iterations. Algorithm \ref{alg:counterexample-detector} provides an overview of the counterexample generator.

\vskip 0.1em
\section{Statistically Measuring Differential Privacy}

\subsection{Adapting PyStatDP to ML Tasks}

We have extended the privacy violation detection framework StatDP \cite{ding_detecting_2018}, to apply it to machine learning pipelines (it has only been used for statistical methods so far). In particular, we have contributed to a fork of that repository, PyStatDP \cite{openmined_github_2020}.

When applying PyStatDP to machine learning, the concepts as described in Section \ref{statdp} will represent the following:

\vspace{-0.3cm}
\begin{itemize}
    \item The set of adjacent databases $(D, D_x)$ will represent training datasets of features and their respective targets that differ by the exclusion of one datapoint (ie. $D$ comprises the full dataset, while $D_x$ excludes row x).
    \vspace{-0.2cm}
    \item The private mechanism $\mathcal{M}$ will represent: 1. training a private model on the dataset $D$ and 2. using the model to predict for a given test input $args$. We can think of $\mathcal{M}(D) = predict(model(D), args)$.
    \vspace{-0.2cm}
    \item The event E will represent the prediction output of our privately trained model $\mathcal{M}$, where this could be a single categorical or regressive output for a single input test datapoint, or a list of outputs for a batch of test datapoints.
\end{itemize}
\vspace{-0.2cm}

Algorithm \ref{alg:test-mechanism} outlines our modified test mechanism $M$ that is adapted to the machine learning context.

\subsubsection{Database Generation}

The counterexample generator first generates a list of candidate tuples of the form $(D_1, D_2, args)$ where $D_1 \sim D_2$. These tuples are likely to form the basis of counterexamples that illustrate violations of differential privacy. Following our modification to the pipeline as described above, the inputs to the counterexample detector are now the indices of rows in the dataset, where an input of $x$ would instruct the mechanism to train a model on the dataset $D_x$.

\subsubsection{Event Selection}

The EventSelector finds events E that are most likely to show violations of $\epsilon$-differential privacy. In the machine learning context, the event $E$ represents the prediction output of our model for a certain test input $args$. Since we are trying to find a statistical difference between the outputs for models trained on adjacent datasets $D$ and $D'$, we want the output E to have as much variation as possible, such that subtle variations of the same model would lead to different outcomes for a given input $args$. 

We do this in two ways: first, finding suitable test data points $args$ that lie close to the decision boundary for classification tasks. Small variations in the training process could shift the decision boundary in ways that lead to different prediction outcomes for the same data point. Second, extending $args$ to be a batch of points such that the model output $E$ becomes a list of predicted labels/targets. This would lead to more possible variations in the model output, without having to search as extensively for a suitable data point.

\subsubsection{Extension to Group Privacy}

As datasets for most machine learning tasks are too large to search through each possible adjacent dataset, we extend the framework to cover the notion of group privacy. We define the adjacent datasets here to differ by $k$ data points, so the resultant privacy measured by the framework will be $k\epsilon$.

\vspace{-0.15cm}
\begin{algorithm}[h]
   \caption{Overview of Test Mechanism}
   \label{alg:test-mechanism}
\begin{algorithmic}
   \STATE {\bfseries Input:} privacy for model training $\epsilon_0$, adjacent dataset $x$, test input features $args$, dataset ($\mathcal{X}$, $\mathcal{Y}$)
   
   \STATE // Construct adjacent dataset
   \IF{$x \neq -1$}
   \STATE $X\_train \longleftarrow \mathcal{X} \setminus x$
   \STATE $y\_train \longleftarrow \mathcal{Y} \setminus x$
   \ELSE
   \STATE // Train using the full dataset
   \STATE $X\_train \longleftarrow \mathcal{X}$
   \STATE $y\_train \longleftarrow \mathcal{Y}$
   \ENDIF
   
   \STATE // Train $\epsilon$-DP model
   \STATE $Model \longleftarrow TrainModel(\epsilon_0, X\_train, y\_train)$
   \STATE $X\_test \longleftarrow \mathcal{X}[args]$
   \STATE $Preds \longleftarrow Predict(Model, X\_test)$
   
   \textbf{return} Preds
   
\end{algorithmic}
\end{algorithm}

\vspace{-0.25cm}
\subsection{Interpreting Experimental Results}

\begin{figure*}[h]
     \centering
     \begin{minipage}[t]{0.3\textwidth}
         \includegraphics[width=\textwidth]{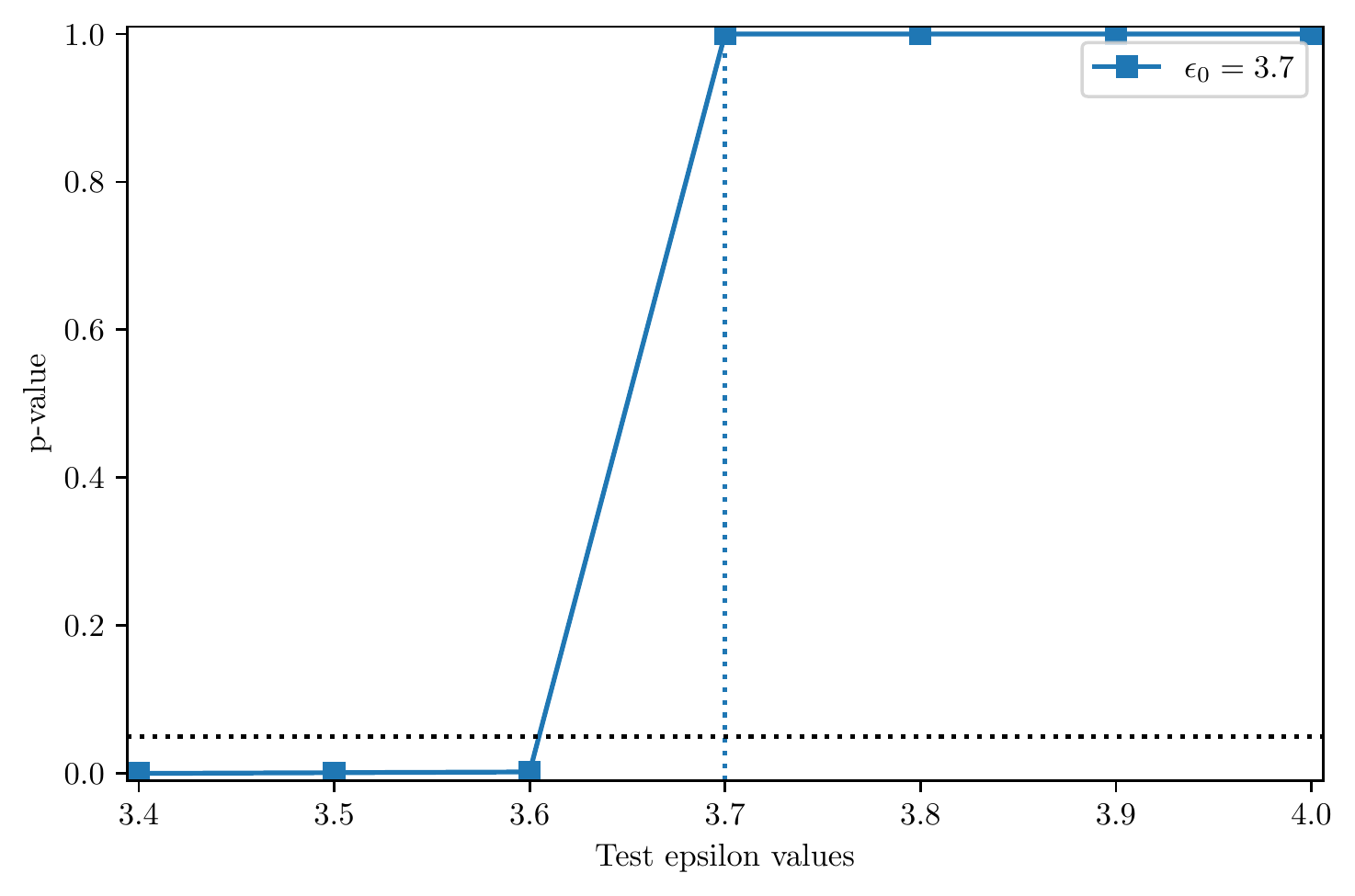}
         \vspace{-0.8cm}
         \caption{Sample output from framework}
         \label{sample-output}
     \end{minipage}
     \hfill
     \begin{minipage}[t]{0.3\textwidth}
         \includegraphics[width=\textwidth]{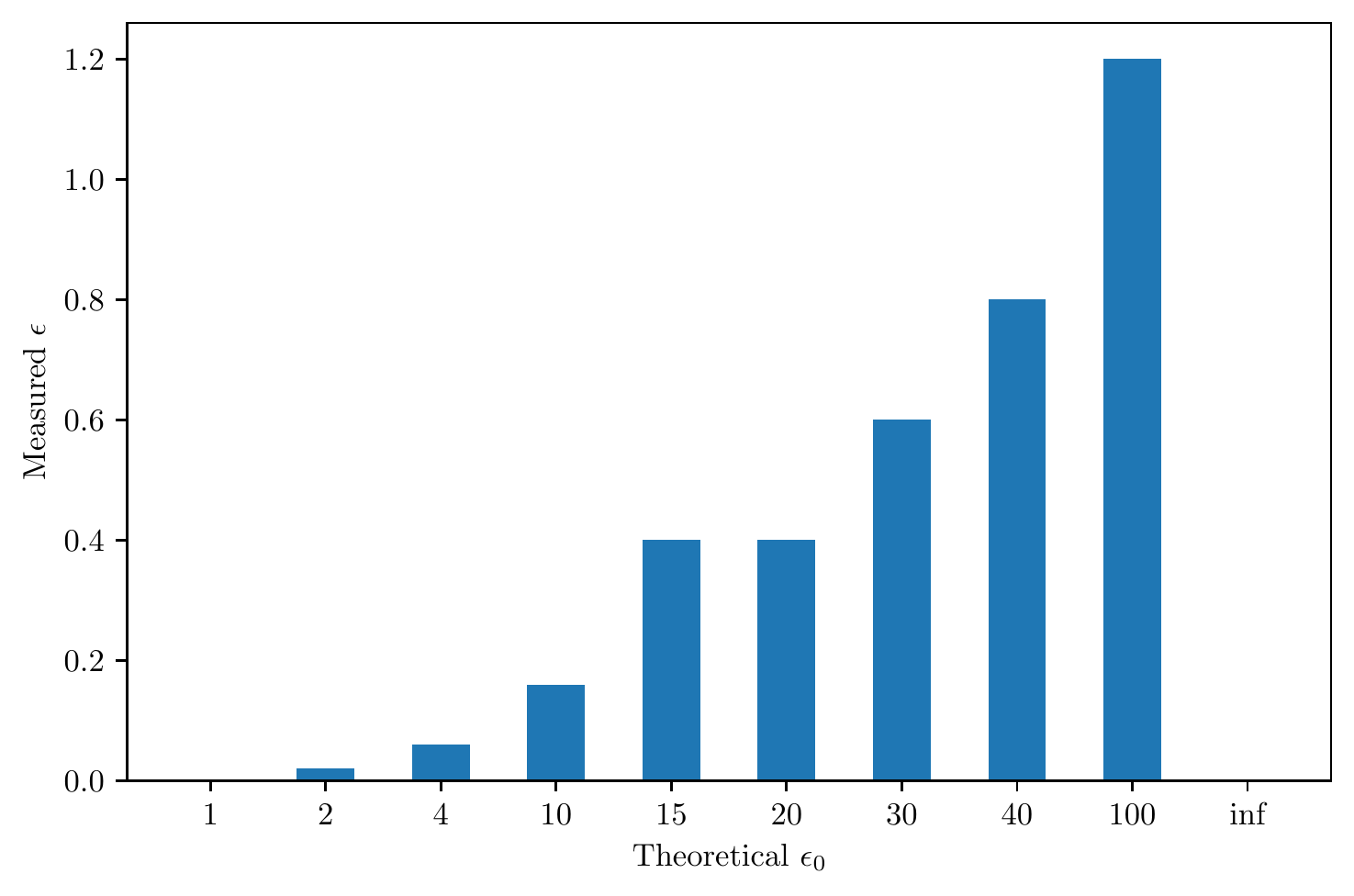}
         \vspace{-0.8cm}
         \caption{Measured $\epsilon$ - Diabetes dataset}
         \label{measured-diabetes}
     \end{minipage}
     \vspace{-0.2cm}
\end{figure*}

Figure \ref{sample-output} shows an example output of the framework. For a given mechanism $\mathcal{M}$ that claims to be $\epsilon_0$-DP, for each $\epsilon$ close to $\epsilon_0$, we test whether $\mathcal{M}$ satisfies $\epsilon$-DP. Each hypothesis test returns a $p$-value for the given $\epsilon$, and the resultant list of $p$-values are plotted in a graph as shown.

We will take the significance level $\alpha$ to be 0.05. $p < \alpha$ indicates that there is less than a 5\% probability that $P(M(D) \in E) \leq  e^\epsilon * P(M(D') \in E)$, so we can reject the null hypothesis that the mechanism $M$ satisfies $\epsilon$-DP (ie. the mechanism does not satisfy $\epsilon$-DP). The converse indicates that the algorithm probably satisfies $\epsilon$-DP. By running the hypothesis test over a range of $\epsilon$ values, we can get an estimate of the actual level of privacy provided by the mechanism, by taking it as the first $\epsilon$ value whose $p$-value rises above $\alpha$. Figure \ref{sample-output} shows an algorithm that correctly satisfies the claimed $\epsilon_0=3.7$ differential privacy.


\vskip 0.1em
\section{Experiments}


We perform our experiments with differentially private models from IBM’s \texttt{diffprivlib} library \cite{holohan_diffprivlib_2019}, and resampling methods from \texttt{imblearn} \cite{lemaitre_imbalanced-learn_2017}. For experiments to validate our technique, we use Gaussian Naive Bayes on the Iris dataset, and Linear Regression on the Adult, Diabetes and Machine CPU datasets \cite{dua_uci_2019}. 

We note that model accuracy is an unnecessary consideration in our experiments, as our framework is solely concerned in the variability in model output as we will demonstrate in the next section.

\subsection{Validation of our Detection Framework}

\begin{figure*}[h!]
     \centering
     \begin{minipage}[t]{0.3\textwidth}
         \includegraphics[width=\textwidth]{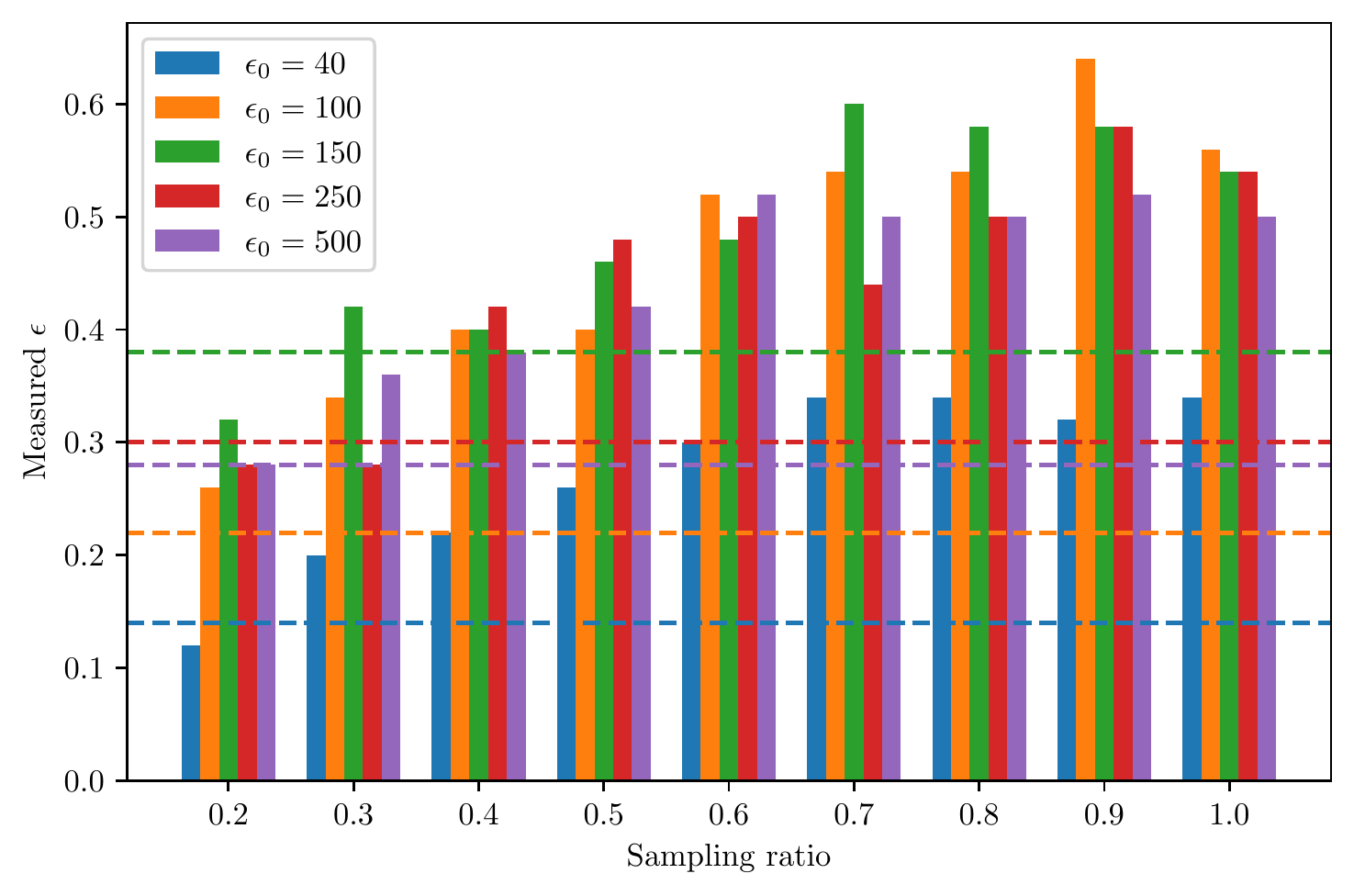}
         \vspace{-0.8cm}
         \caption{Measured $\epsilon$ from oversampling - Machine CPU dataset}
         \label{oversampling-machine}
     \end{minipage}
     \hfill
     \begin{minipage}[t]{0.3\textwidth}
         \includegraphics[width=\textwidth]{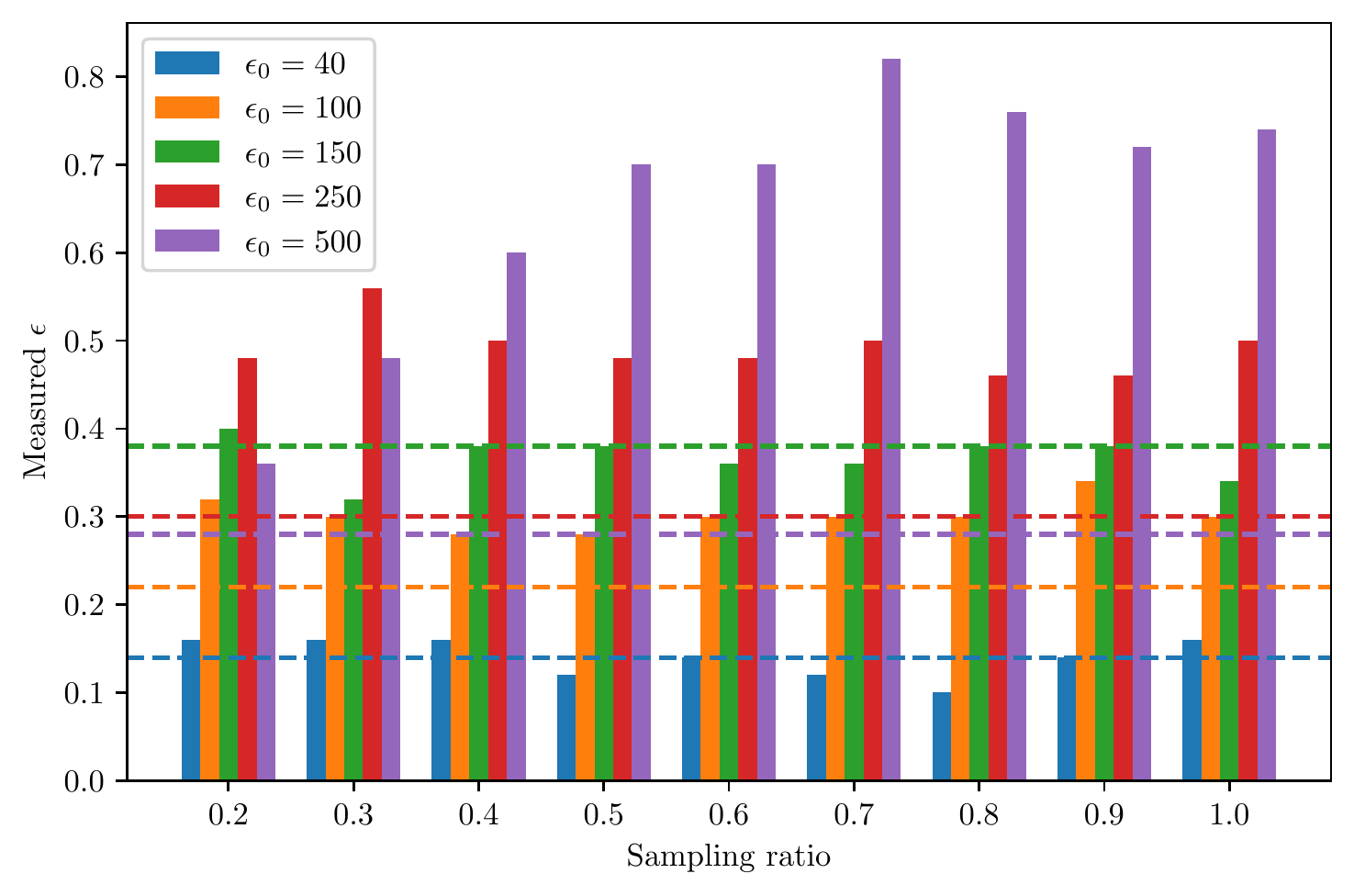}
         \vspace{-0.8cm}
         \caption{Measured $\epsilon$ from subsampling - Machine CPU dataset}
         \label{subsampling-machine}
     \end{minipage}
      \hfill
     \begin{minipage}[t]{0.3\textwidth}
         \includegraphics[width=\textwidth]{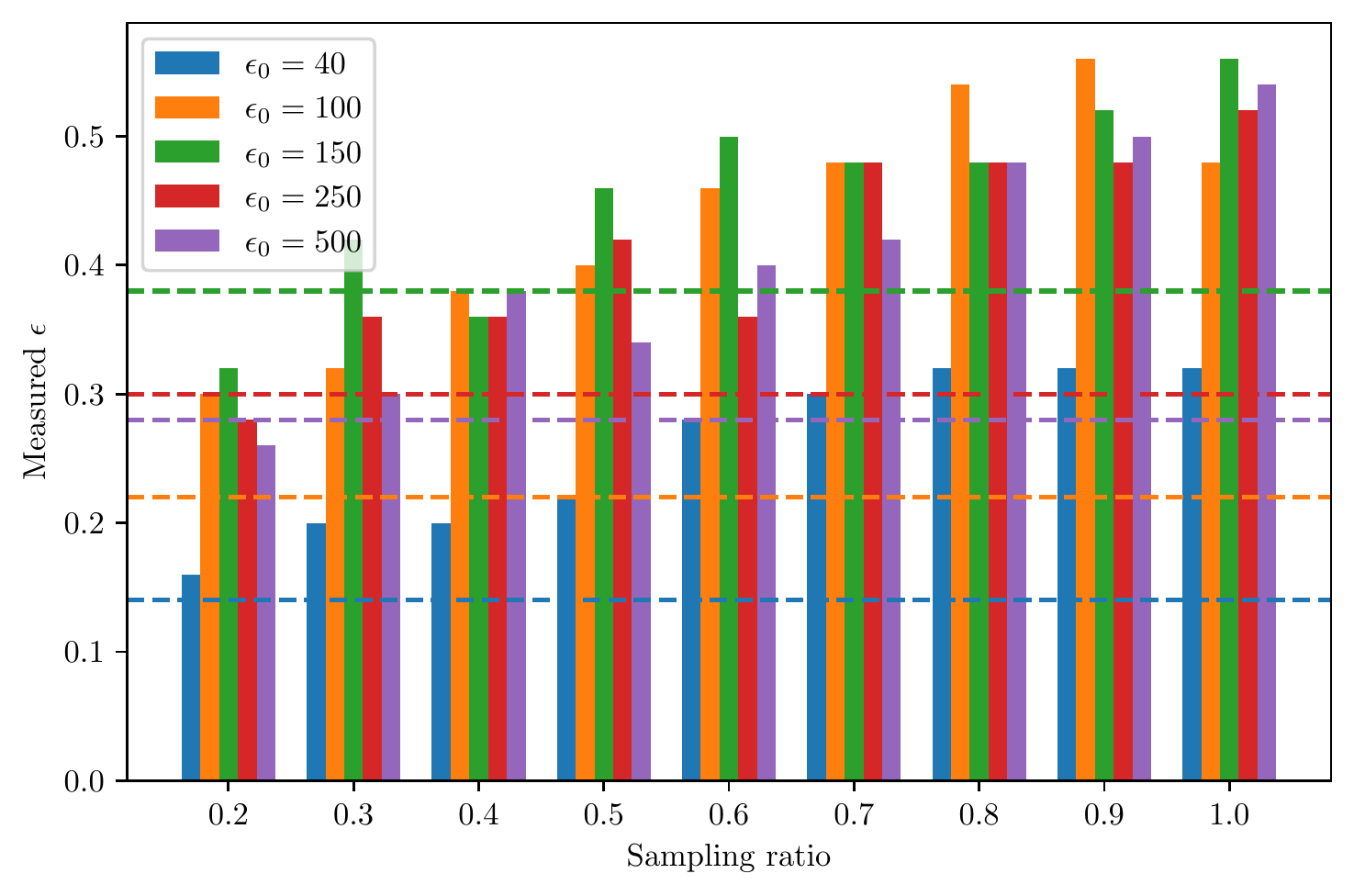}
         \vspace{-0.8cm}
         \caption{Measured $\epsilon$ from SMOTE resampling - Machine CPU dataset}
         \label{smote-machine}
     \end{minipage}
    \vspace{-0.2cm}
\end{figure*}

Figure \ref{measured-diabetes} shows the measured $\epsilon$ values when using our framework to test the DP Linear Regression model on the Diabetes dataset. For each level of $\epsilon_0$, the R2-score of the model stays constant at $\sim\hspace{-0.1cm}0.03$, while the framework measures a higher $\epsilon$. These results show the correctness of our framework as we can observe that it detects variability in our model outputs due to the level of training $\epsilon$, and not due to the model accuracy. We also see that the non-private model achieves $\epsilon=0$, which may be because the model has generalised so well that its output does not change when exposed to more samples. These results have been reproduced with other models and datasets.

One limitation is that we are only able to measure $\epsilon$ at high levels of $\epsilon_0\gg1$. As we have high compute requirements of having to run enough iterations to achieve statistical significance, we only use model predictions of a few data points to distinguish between the adjacent datasets. This is comparable to, but weaker, than the black-box adversary used in \cite{nasr_adversary_2021}, which achieved a lower bound of $\sim\hspace{-0.15cm}10\%$ of the theoretical $\epsilon_0$. Hence achieving a lower bound of $\sim\hspace{-0.1cm}0.5\%$ on predictions from just $\sim\hspace{-0.1cm}3$ data points is reasonable given the relative ease of using our framework.

\subsection{Effect of Resampling on Privacy}

As our framework has greater measurement sensitivity to regression outputs, we perform our resampling experiments on imbalanced regression tasks, using a Linear Regression model. Here, we classify ‘outlier’ points in a one-tailed manner, thresholding the minority/majority class using a 10/90 split.




Figures \ref{oversampling-machine}, \ref{subsampling-machine} and \ref{smote-machine} show measured epsilon values from performing various levels of resampling as a preprocessing step on the imbalanced Machine CPU dataset. The dashed lines represent baseline $\epsilon$ values without sampling.

\subsubsection{Oversampling}

We can see that for the oversampling techniques, increasing the resampling ratio corresponds with a higher level of $\epsilon$ measured on the resultant model. This confirms our hypothesis that resampling increases the privacy leakage of the resultant model. We note that SMOTE \cite{chawla_smote_2002} resampling, which generates synthetic data points, causes a relative lower increase in privacy leakage.

\subsubsection{A Note on Dataset Size}

We should be cautious to take these results at face value, because an increase in the level of oversampling increases the size of the resultant dataset, and vice versa for subsampling. With a constant training $\epsilon_0$, a larger dataset results in less noise required to be added to the training process, which may exert an upward force on the measured privacy for oversampling (and vice versa for subsampling).

At higher values of $\epsilon_0$, this effect of dataset size should be negligible, since very little noise is already being added. Since we can see that even at $
\epsilon_0=500$ the trend still continues upward, this validates our hypothesis on the effect of resampling on privacy leakage.

\subsubsection{Subsampling}

Results from our subsampling experiment in Figure \ref{subsampling-machine} seem to prove our hypothesis further. With an increase in subsampling ratio, the measured $\epsilon$ remains relatively constant, apart from when $\epsilon_0=500$.

Assuming the effects of the dataset size as discussed above are present on the resultant amount of noise added to the model training, the fact that the measured $\epsilon$ remains relatively constant suggests that the increased privacy leakage caused by subsampling balanced out the decrease in privacy leakage from the greater amount of noise added to the training. The trend seen with $\epsilon_0=500$ validates this, since at this high level of $\epsilon_0$ so little noise is added that the effect from the decreased dataset size is negligible. Hence the increase in measured $\epsilon$ can be attributed to the increased privacy leakage due to the subsampling itself.

\vskip 0.1em
\section{Conclusions and Future Work}

We have presented a novel way of measuring privacy levels of machine learning algorithms through a statistical approach. While this may not achieve a tight lower bound, it serves as a simpler tool to measure and compare DPML pipelines as compared to typical approaches.

Through our experiments we have also shown that typical resampling techniques used to deal with imbalanced datasets leak significant amounts of privacy. Future work includes exploring techniques that generate synthetic datasets \cite{branco_smogn_2017, gretel_gretel_2021}, other augmentation methods and experiments on deeper models.


\end{document}